\documentclass[conference]{IEEEtran}

\usepackage{cite}
\usepackage{graphicx}
\usepackage{booktabs}
\usepackage{array}
\usepackage{url}
\usepackage{xcolor}
\usepackage{enumitem}
\usepackage{amsmath}
\usepackage{amssymb}
\usepackage{amsthm}
\usepackage{verbatim} 
\usepackage{subcaption}
\usepackage{comment}
\usepackage{xspace}
\usepackage{hyperref}
\usepackage[absolute,overlay]{textpos}

\newcommand{\eg}{\textit{e.g.},\xspace} 

\begin{document}

\title{Beyond the Model: The Critical Role of Data Filtering in Clinical Machine Learning}


  \author{\IEEEauthorblockN{
  Noah Subedar \IEEEauthorrefmark{1}\IEEEauthorrefmark{5},
  Colin Campbell \IEEEauthorrefmark{1}\IEEEauthorrefmark{5},
 Wenjing Zhang\IEEEauthorrefmark{1},
  Dan Perri\IEEEauthorrefmark{2}\IEEEauthorrefmark{3}\IEEEauthorrefmark{4},
  Sarah Culgin\IEEEauthorrefmark{2}\IEEEauthorrefmark{3},
  and Andrew Hamilton-Wright\IEEEauthorrefmark{1}
  }
  \IEEEauthorblockA{\IEEEauthorrefmark{1}School of Computer Science,
  University of Guelph, Ontario, Canada
  ~~ Email: wzhang25@uoguelph.ca}
  \IEEEauthorblockA{\IEEEauthorrefmark{2}St. Joseph's Healthcare Hamilton,
  Hamilton, Ontario, Canada}
  \IEEEauthorblockA{\IEEEauthorrefmark{3}The Research Institute of St. Joe's Hamilton,
  Hamilton, Ontario, Canada}
  \IEEEauthorblockA{\IEEEauthorrefmark{4}McMaster Univeristy,
  Hamilton, Ontario, Canada}
  \IEEEauthorblockA{\IEEEauthorrefmark{5}Equal contribution and co-first authors.}
}

\maketitle

\vspace{-50px}
\begin{abstract}
Machine learning (ML) studies using clinical data often rely on preprocessing and filtering pipelines before model development. The filtering decisions made in these pipelines can alter the dataset's statistical structure and may artificially reduce or increase the complexity of the prediction task. We argue that filtering choices should be treated as part of the scientific method rather than as a routine preprocessing step. We further discuss the need for explainable and transparent preprocessing pipelines that allow researchers to understand why specific filtering choices are made and how these choices affect the resulting data distribution and model performance. All of the source code for this work is available on \color{blue}\href{https://github.com/WenjingZhang-Guelph/AIML_Systems_2026}{GitHub}. 
\end{abstract}


\begin{textblock*}{7.5in}(0.25in,0.25in)
\textcolor{teal}{Presented at the 6th International Conference on AI-ML Systems,
Politecnico di Milano Campus in Lecco, Italy
Oct 9, 2026}\\
\centering\url{https://www.aimlsystems.org/2026/}
\end{textblock*}
\vspace{-10pt}
\section{Introduction}
\label{sec:introduction}

Machine learning has become an increasingly central tool in modern clinical medicine, with applications spanning diagnosis, risk stratification, and patient outcome prediction across nearly every specialty \cite{topol2019high, ghosheh2024survey}. Driven by the growing availability of large-scale electronic health records (EHRs) and clinical databases, predictive models now hold real potential to support clinical decision-making, reduce diagnostic error, and improve the efficiency of care delivery \cite{wang2024recent}. Realizing this potential, however, depends critically on how the underlying clinical data is prepared. Clinical ML research often begins with substantial preprocessing. Missing values, outliers, long gaps in patient records, irregular measurements, and physiologically implausible observations are commonly removed or transformed before training. While these steps are often necessary, they can also change the underlying prediction problem. A central concern in this work is that filtering may remove difficult cases, reduce variance, separate class distributions, or otherwise alter the data in ways that improve apparent model performance without necessarily improving clinical usefulness. This issue is especially important when models are evaluated on published datasets that require cleaning or other preprocessing, where researchers may compare algorithms without fully understanding how preprocessing decisions shaped the data.

Our motivation comes from three practical realities of working with clinical data, which we illustrate through the lens of decision-making on real-world Intensive Care Unit (ICU) datasets \cite{johnson2023mimic, johnson2016mimic, pollard2018eicu}. Our first motivation is that filtering can change task difficulty: it can alter class separation, remove complex cases, and change the distribution of input features. For example, if filtering removes records with long measurement gaps or high variability, the remaining data may be easier for a model to classify, particularly if decision regions with conflicting nearby classifications are removed. Our second motivation is that different filters may affect different tasks differently. A filtering rule that helps one task may hurt another; ICU transfer prediction, mortality prediction, and length of stay prediction may all respond differently to the same filtering strategy. Our third motivation is that preprocessing pipelines need to be explainable. Providing code or cutoff tables alone is not enough, as researchers also need to explain why each filtering decision is appropriate for the clinical and biological problem being studied if the result is intended for application.

Our main contribution is demonstrating that data filtering decisions can substantially alter the difficulty of ML tasks and, consequently, influence the validity and comparability of experimental results. We argue that filtering choices should be treated as part of the scientific method rather than as a routine preprocessing step, and that they should be explicitly reported, justified, and analyzed as an integral part of the experimental design. To support this claim, we developed an explainable data filtering toolbox and used it to systematically investigate the effects of different filtering strategies on ML tasks. We further compared these effects with those produced by the widely adopted MIMIC-Extract preprocessing pipeline \cite{wang2020mimic}.

Concretely, this paper addresses the following  questions:
\begin{enumerate}[leftmargin=0.3cm]
    \item How do different filtering strategies affect the statistical
    structure of clinical datasets?
    \item Which filtering choices make downstream ML tasks
    easier or harder?
    \item How can researchers detect when preprocessing has changed the
    meaning or difficulty of a prediction task?
    \item How can one develop an open source visualization tool to support explainable
    and transparent preprocessing pipelines?
\end{enumerate}
Characterizing these statistical shifts is necessary because missingness in EHRs is described as widespread and inescapable, making it a defining property of those datasets rather than edge cases \cite{shi2024}. Existing evidence rejects the assumption of independence of missingness patterns in certain clinical settings, demonstrating that missing data in clinical practice is likely not random \cite{suzen2023}. EHR data gathering must therefore be treated as a joint product of underlying physiology and clinician decision-making \cite{mulyadi2019}. To address these questions, our contributions are fourfold:
\begin{enumerate}[leftmargin=0.3cm]
   
   \item {Empirical evidence that filtering reshapes the data and the task.} We systematically apply eleven filtering strategies to three publicly available ICU databases \cite{johnson2023mimic, johnson2016mimic, pollard2018eicu} and show that filtering decisions measurably alter the statistical structure of the cohort, shifting class centroids, changing feature distributions, and removing clinically meaningful edge cases, rather than acting as neutral data cleaning.

    \item {Demonstration that filtering can make tasks artificially easier or harder.} Using a  Random Forest evaluation protocol across multiple aggregation methods and random seeds, we quantify how each filter changes downstream performance on two binary prediction tasks (prolonged ICU stay and in-hospital mortality), showing that the same filter can improve one task while degrading another.

    \item {A diagnostic methodology for detecting problem-altering preprocessing.} We introduce a practical workflow, combining per-class centroid deviation analysis with McNemar significance testing against an unfiltered baseline, that allows researchers to distinguish filters that legitimately remove artifacts from those that fundamentally change the prediction problem.

    \item {An open source and explainable data filtering toolbox available on \color{blue}\href{https://github.com/WenjingZhang-Guelph/AIML_Systems_2026}{GitHub}.} We develop and release a toolbox that lets researchers apply, visualize, and compare filtering strategies while tracking their impact on the resulting dataset, supporting transparent and reproducible preprocessing pipelines.
\end{enumerate}

\section{Related Work}

\subsection{Datasets for Healthcare}
The development of robust ML models for healthcare, particularly in ICUs, is frequently hindered by the scarcity of high-quality datasets and strict privacy regulations that create institutional data silos. The scarcity of data sets leads to the reuse of available data sets for purposes other than those anticipated by the original collection team. This invites filtering and cleaning of the data to remove outliers or collection artifacts. Our main point is that this must be done with an understanding of the underlying data problem, or the resulting filtered data may no longer fairly represent the underlying interesting clinical problem. This general concern is not new and is noted in textbooks on data collection and analysis \cite{rea_parker_2002surveydesign, pham2023statistics}; what has been missing is a systematic way to detect when a specific filtering choice has crossed from cleaning into distortion, which we discuss in Section~\ref{subsec:eval-gap}.

\vspace{-5pt}
\subsection{How Preprocessing Alters Clinical Data}
Structuring irregular intensive care data for standard models generally requires a preprocessing and filtering pipeline~\cite{wang2020mimic, harutyunyan2019multitask}. To address the sparse and asynchronous nature of continuous physiological events, researchers commonly employ missing-data handling techniques, such as forward-filling, mean imputation, or spline interpolation, to map the data onto fixed temporal grids. Raw electronic health records also contain sensor artifacts, irregular sampling intervals, and extended periods of unmonitored activity, which standard pipelines typically address through outlier removal, physiological plausibility checks, and time-gap filtering. While these practices produce the mathematically tractable, complete datasets that many models expect, they can introduce a trade-off. The timing and presence of clinical measurements are themselves shaped by care processes and may carry predictive signal \cite{agniel2018biases, che2018recurrent}, so forcing this information onto regular grids risks discarding informative temporal structure or introducing imputation artifacts; the choice of imputation method itself can affect downstream model performance, and methods designed for statistical inference are not always optimal for prediction \cite{digitale2025methods}. Filtering that removes highly irregular trajectories or flattens variance can shift class distributions and remove clinically meaningful edge cases, causing the dataset to diverge from the clinical setting it represents and adding to the reproducibility challenges identified in machine learning for health~\cite{mcdermott2021reproducibility}.

\vspace{-5pt}
\subsection{Preprocessing in Clinical ML Benchmarks and the Evaluation Gap}
\label{subsec:eval-gap}

Despite these well-understood trade-offs, clinical ML research commonly evaluates predictive models on benchmark datasets such as MIMIC-III \cite{johnson2016mimic}, MIMIC-IV \cite{johnson2023mimic}, and eICU \cite{pollard2018eicu}, using tasks including mortality prediction, length of stay prediction, and patient outcome forecasting. Existing studies primarily focus on model development and performance comparison, typically reporting metrics such as AUROC, AUPRC, and F1-score. To facilitate experimentation, standardized extraction pipelines such as MIMIC-Extract \cite{wang2020mimic} and benchmark suites \cite{harutyunyan2019multitask,mesinovic2025survbench} fix a common set of preprocessing choices to enable fair comparison. However, such pipelines focus on standardization for reproducible comparison and do not characterize how their preprocessing choices affect the difficulty of the resulting tasks. This omission matters because preprocessing is not a neutral operation. To our knowledge, no prior work systematically isolates individual filtering strategies, quantifies how each reshapes the data distribution and downstream task difficulty, and provides a diagnostic for detecting when preprocessing has changed the underlying problem. This work addresses that gap by systematically examining the impact of filtering choices on clinical machine learning evaluation.

\vspace{-5pt}
\section{Methodology}

\vspace{-5pt}
\subsection{Dataset}

We have made our pipeline compatible with three widely used, publicly available ICU databases: MIMIC-III~\cite{johnson2016mimic}, MIMIC-IV~\cite{johnson2023mimic}, and eICU~\cite{pollard2018eicu}. All three contain de-identified electronic health records (EHR) from U.S. intensive care units and are accessed under institutional data use agreements. We focus primarily on MIMIC-III, as it underlies the MIMIC-Extract pipeline~\cite{wang2020mimic} that we use as our reference baseline.

\subsubsection*{Cohort Definition}

For each of our datasets, we extract all hospital admissions for which the vital sign data is available within the first 24 hours of admission. Admissions are excluded if label data (ICU stay duration or mortality) could not be linked to the charted vital signs. We then represent each valid admission as a multivarible time series spanning $T = 24$ hourly intervals and $V = 7$ vital sign channels. This gives us a raw observation matrix $\mathbf{X} \in  \mathbb{R}^{24\times7}$. In its pre-aggregated form, each cell $(t, v)$ contains the measurement recorded for vital $v$ during hour $t$ of the admission; the number of measurements per cell is variable and may be zero.
  
\subsubsection*{Variables}

The seven vital signs that are a part of our pipeline can be seen in Table~\ref{tab:vitals}. As a part of our preprocessing, we also converted all of the temperature data to degrees Celsius using the standard  $T_{\text{C}} = \frac{T_{\text{F}} - 32}{1.8}$ formula. 

For each of these vital signs, there is a range that we determined by considering what the metric measures, as well as some loose physiological constraints that we can place on each metric. All of these values are available in Table~\ref{tab:vitals} and will be discussed in more detail in section \ref{subsec:filt}.

\begin{table}[t]
    \centering
    \caption{Vital sign features and physiologically validated ranges used for outlier filtering.}
    \label{tab:vitals}
    \begin{tabular}{llll}
        \hline
        \textbf{Feature} & \textbf{Abbreviation} & \textbf{Valid Range} & \textbf{Unit} \\
        \hline
        Heart Rate & HR & $[1, 600]$ & bpm \\
        Systolic Blood Pressure & SBP & $[1, 400]$ & mmHg \\
        Diastolic Blood Pressure & DBP & $[1, 300]$ & mmHg \\
        Mean Blood Pressure & MBP & $[1, 300]$ & mmHg \\
        Respiration Rate & RR & $[1, 70]$ & breaths/min \\
        Temperature & Temp & $[21, 50]$ & °C \\
        Oxygen Saturation & SpO$_2$ & $[1, 100]$ & \% \\
        \hline
    \end{tabular}
    \vspace{-15pt}
\end{table}

\subsubsection*{Clinical Tasks}
While these datasets support many possible downstream tasks, we focus on two independent binary prediction tasks defined for each admission. The \textbf{Prolonged ICU Stay (ICU)} label is positive if the total ICU time across all episodes within the admission exceeded three days, and negative otherwise. The \textbf{Mortality} label is positive for a record if the patient became deceased during their hospital stay, and negative if they survived to discharge.

\vspace{-5pt}
\subsection{Filtering Strategies}\label{subsec:filt}

We examine eleven filtering strategies applied to the raw vital data. Filters are partitioned into two stages:\ \textit{pre-aggregation} filters that operate on individual measurements within each list-valued cell and \textit{post-aggregation} filters that operate on the matrix produced after hourly aggregation. Each filter is applied independently to the unfiltered dataset, producing a separate filtered dataset for each strategy; filters are not applied sequentially except where explicitly noted. Below is a description of what each of our individual filters is, as well as the thought process that went into each of them. For a quick overview, Table~\ref{tab:filters} summarizes all of the filtering strategies and their key parameters.

\begin{description}[leftmargin=0.3cm]

\item[Outlier filtering (individual vitals):] We have seven different filters that each remove physiologically implausible measurements from each cell before aggregation. For a vital sign with valid range $[L, U]$, any individual measurement $x_i$ satisfying $x_i < L$ or $x_i > U$ is removed from its hourly list:
\begin{equation}x_i \;\leftarrow\; \begin{cases} \text{removed} & x_i < L \;\text{ or }\; x_i > U \\ x_i & \text{otherwise} \end{cases}\end{equation}

The bounds in Table~\ref{tab:vitals} are not intended to remove valid data that resulted from an event that occurred, but rather remove values that clearly make no sense and are likely the result of a sensor error or artifact in the underlying data. For example, during Ventricular Fibrillation, an adult's heart rate can reach up to 500 beats per minute \cite{burns_buttner_2024_vf}, and based on our heart rate filter, we would not omit this value as it is physically possible and very clinically relevant to any downstream task to preserve this data. However, if we were to encounter a negative heart rate or a heart rate of several thousand beats per minute, there is no way for a human heart to reach these levels. Therefore, there must be something wrong with our underlying data, which is why it is being filtered. Hourly cells whose lists become empty after filtering are set to \texttt{NaN}. The seven individual vital filters (HR, SBP, DBP, MBP, RR, Temp, SpO$_2$) are each evaluated separately, and their union is evaluated as the \textbf{All Vitals} composite filter.

\item[Missing data imputation:] Applied post-aggregation, this filter imputes \texttt{NaN} cells in each vital sign column using $k$-nearest-neighbour (KNN) regression with $k = 3$ neighbours, operating independently along the time axis. Imputation is applied only when at least two known values exist in the column; otherwise, the column is left as-is. This filter operates at the feature level, meaning it modifies cell values but will never remove records.

The imputed value at the missing time index $i$ is the distance-weighted mean of the three temporally closest observed values:
{%
\begin{equation}
\hat{x}_i = \frac{\sum_{j \in \mathcal{K}} |i - j|^{-1} \cdot x_j}{\sum_{j \in \mathcal{K}} |i - j|^{-1}}, \quad |\mathcal{K}| = 3
\end{equation}
}%


This strategy reflects the observation that ICU documentation is episodic rather than continuous; short gaps between recorded values are more likely to represent charting intervals than true physiological absences.

\item[Long missing segment filtering:] A patient record is removed if any single vital sign column contains a run of 10 or more consecutive missing or zero values. A 10-hour gap in a single vital sign column most likely indicates the patient was disconnected from that specific monitor for an extended period. Gaps of this length are too long to bridge reliably with imputation and would introduce a large block of estimated data into an otherwise observed record.

\item[Long gap filtering:] A record is removed if two or more consecutive hourly rows are empty. That is, all seven vital signs are simultaneously missing or zero over this time period. A simultaneous blackout across all modalities is a stronger signal of systemic monitoring failure (\eg the patient was transferred to the operating room, resulting in equipment disconnection) than a single-vital gap, and the threshold is correspondingly stricter.

\item[High invalid data filtering:] A record is removed if the fraction of invalid cells (NaN or zero) across the entire $24 \times 7$ matrix exceeds 10\% as seen in Equation~\ref{high_invalid_data_equ}. This filter catches broadly sparse records that lack any single long gap but whose overall data density is insufficient for reliable feature extraction. The 10\% threshold is conservative by design, retaining records that miss up to 16 of 168 possible cell values.
\vspace{-0.5cm}
\begin{equation}r = \frac{\displaystyle\sum_{t,v}\, \mathbf{1}[x_{t,v} \in \{0, \texttt{NaN}\}]}{24 \times 7} \;\geq\; 0.10\label{high_invalid_data_equ}\end{equation}

\end{description}

\begin{table}
    \centering
    \caption{Summary of filtering strategies}
    \label{tab:filters}
    \begin{tabular}{llll}
        \hline
        \textbf{Filter} & \clap{\begin{tabular}{@{}c@{}}\textbf{Aggregation}\\\textbf{Stage}\end{tabular}} & \textbf{Level} & \textbf{Threshold} \\
        \hline
        Heart Rate & Pre & Value & $[1, 600]$ bpm \\
        Systolic BP & Pre & Value & $[1, 400]$ mmHg \\
        Diastolic BP & Pre & Value & $[1, 300]$ mmHg \\
        Mean BP & Pre & Value & $[1, 300]$ mmHg \\
        Respiration Rate & Pre & Value & $[1, 70]$ breaths/min \\
        Temperature & Pre & Value & $[21, 50]$ °C \\
        Oxygen Saturation & Pre & Value & $[1, 100]$ \% \\
        All Vitals & Pre & Value & Union of above \\
        Fill Missing Data & Post & Feature & KNN $k{=}3$, ${\geq}2$ known pts \\
        Long Missing Segment & Post & Record & ${\geq}10$ consec.\ missing/zero \\
        Long Gap & Post & Record & ${\geq}2$ consec.\ empty rows \\
        High Invalid Data & Post & Record & ${\geq}10\%$ invalid cells \\
        \hline
    \end{tabular}
    \vspace{-15pt}
\end{table}

\vspace{-5pt}
\subsection{Motivation for Outlier Filtering}
\begin{figure}[b]
    \vspace{-15pt}
    \centering
    \includegraphics[width=1\linewidth]{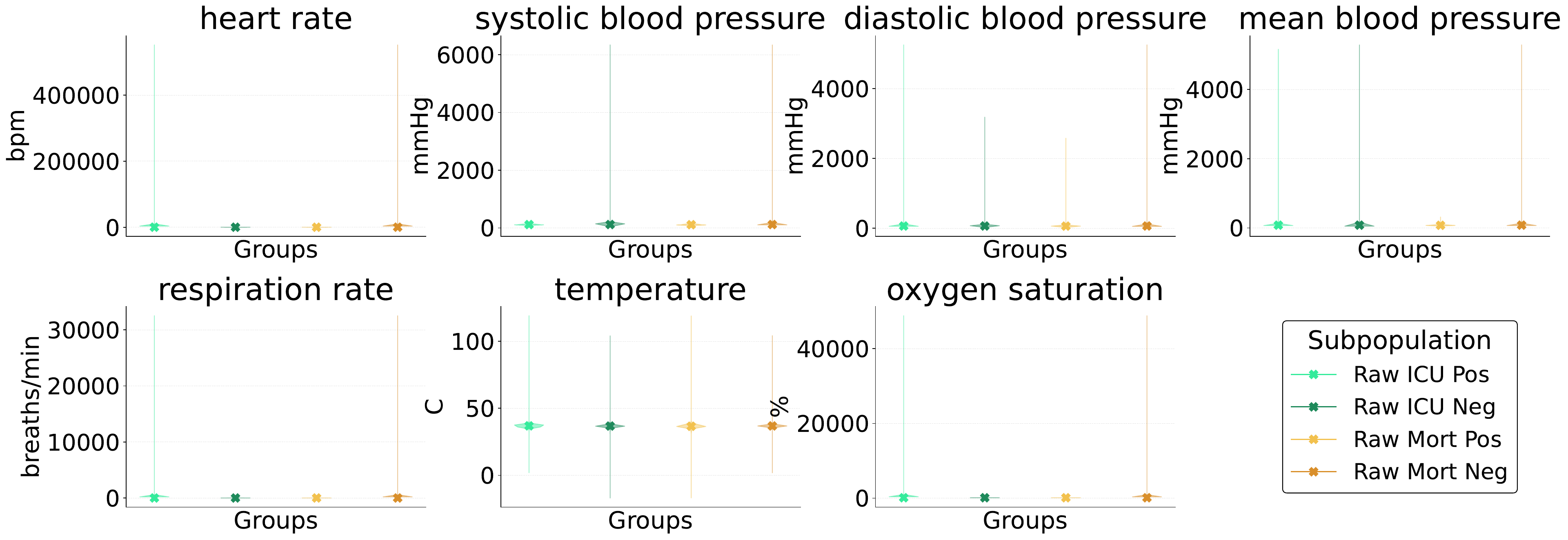}
    \caption{Centroid density of the raw MIMIC-III dataset}
    \label{fig:centdenbase}
    \vspace{-30pt}
\end{figure}

All of these data sets contain records with physiologically impossible data within them. These data points should be removed or adjusted in some manner prior to attempting any downstream tasks with data. Some of the outliers in the dataset are so extreme that they can be seen by looking at Figure \ref{fig:centdenbase}, where we can see that across the vast majority of our vital signs, there are values that make no sense. For example, looking at oxygen saturation, we know that this value is recorded as a percentage. Therefore, any values over 100\% have to be the result of some sort of sensor error or artifact.  While this seems quite obvious, the fact that this is not typically noted in analysis based on these data sets underscores our point of how critical it is to apply this sort of informed analysis of the presented data.

\vspace{-5pt}
\subsection{Model and Evaluation}
\vspace{-2pt}
\subsubsection*{Feature Representation}

Raw vital sign time series are aggregated from list-valued cells to scalar values using one of the following methods: mean, median, standard deviation, mean absolute deviation, and maximum absolute deviation. Each of these methods is independent of the others and is applied to each $(t, v)$ cell. This produces five variants of each dataset, allowing us to assess whether filter effects are consistent across different aggregation methods. After we complete aggregation, each patient record is represented as a flattened $168$-dimensional feature vector $\mathbf{\tilde{x}} \in  \mathbb{R}^{168}$ (24 hours $\times$ 7 vital signs). Missing values (\texttt{NaN}) remaining after aggregation are replaced with zero prior to model input.

\subsubsection*{Model}

Using this vector, we then train a Random Forest classifier with 300 trees for each of our aggregation methods and filter combinations. Random Forest was selected for its robustness to irrelevant features and its ability to handle the mixed-density feature vectors that arise from variable monitoring coverage across patients.

\subsubsection*{Experimental Protocol}

To keep our experiments consistent, we evaluate each dataset using stratified 5-fold cross-validation, with fold assignments drawn once from the unfiltered cohort by admission identifier, so that every filtered variant inherits the same assignments and all variants are scored on the same patients. Within each fold, the held-out $20\%$ is scored, and the remaining $80\%$ is split $7{:}1$ into a training set and an inner validation set used only to select the decision threshold by Youden's $J$~\cite{youden1950index}. In our experiments, we found that using Youden's $J$'s threshold provided similar results to other class balancing techniques such as SMOTE and oversampling~\cite{chawla2002smote}. Class balancing is especially necessary for the mortality task, as the class prevalence is around 10\%. To account for random variation in data splitting and model initialization, the procedure is repeated over four independent partitions and results are reported as the mean and standard deviation across repeats. 

\subsubsection*{Evaluation Metrics}

For the majority of our experiments, we will focus on the following three model evaluation metrics:

\begin{description}[leftmargin=0.3cm]
    \item[Accuracy:] Proportion of correctly classified samples on the held-out test set.
    
    \item[Macro F1:] The unweighted mean of per-class F1 scores, computed to account for potential class imbalance in the ICU and mortality labels.
    
    \item[McNemar's test:] A paired nonparametric test comparing the prediction errors of each filtered model against the unfiltered (raw) baseline. Majority-vote predictions are aggregated across seeds before constructing the contingency table. This test determines whether a given filter produces a statistically significant change in classification behaviour relative to no filtering.
\end{description}

The primary comparison is between each filtered dataset and the unfiltered baseline, evaluated under the same model, aggregation method, and random seeds. This design isolates the marginal effect of each filter on downstream predictive performance.

\vspace{-5pt}
\subsection{Tool Development}

To systematically investigate the effects of data filtering, we developed an open source and explainable data filtering toolbox for clinical datasets. The toolbox enables researchers to apply, visualize, and compare different filtering strategies while tracking their impact on the resulting dataset. Specifically, it provides tools for examining changes in data distributions before and after filtering, identifying records and patient cases removed by individual filtering steps, and quantifying how filtering alters key dataset characteristics. 

By making filtering decisions transparent and interpretable, the toolbox helps researchers understand how preprocessing choices influence downstream ML tasks and supports more reproducible and trustworthy experimental evaluations.

\vspace{-3pt}
\section{Results and Discussion}
\vspace{-2pt}
\subsection{Filtering Changes the Data Distribution}

\begin{figure}[b]
    \vspace{-20pt}
    \centering
    \includegraphics[width=\columnwidth]{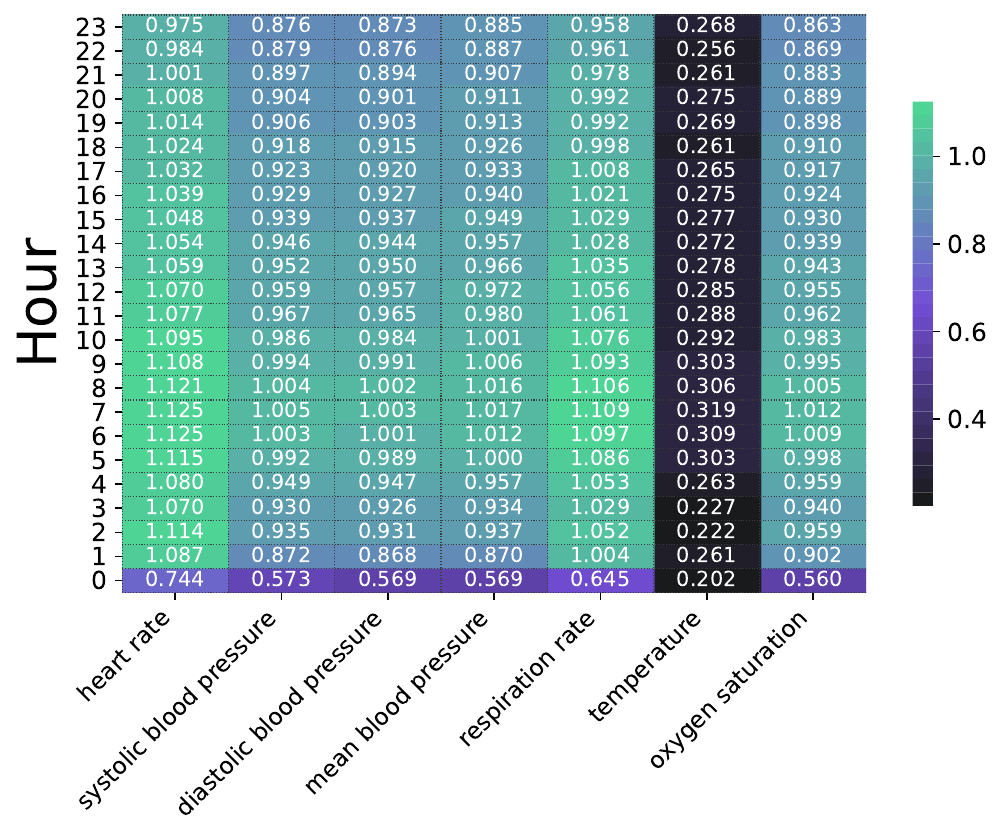}
    \caption{Heatmap of mean hourly observation counts per vital sign across MIMIC-III ICU stays (unfiltered, raw).}
    \label{fig:length_heatmap}
    \vspace{-30pt}
\end{figure}

\looseness=-1 Preprocessing filters are commonly framed as data cleaning steps that remove noise while maintaining the integrity of the original data. However, any filter that selectively discards or modifies observations based on their values will, by definition, alter the statistical properties of the output dataset. We found this to be true across all different levels of filters that we examined in this work. Before any filtering is applied, the raw MIMIC-III timeseries data contains a wide variety of observation densities. As shown in Figure~\ref{fig:length_heatmap}, the mean number of raw measurements per vital-sign/hour cell varies considerably across both the temporal axis and the vital-sign axis, with some combinations receiving far more measurements per hour than others. 

This non-uniformity means that value-level filters do not affect all vitals equally. A filter that removes a fixed number of observations from a densely measured vital will eliminate more absolute data points than the same filter applied to a sparsely measured one. Likewise, if a vital contains fewer overall measurements, then removing data associated with this vital should be done with more caution than a vital with more data points. For example, a vital sign like heart rate can be monitored very easily and frequently, which translates into more data points present on average in records. On the other hand, if we were dealing with something like blood hemoglobin concentration, which requires a blood test and lab work we would see far fewer measurements recorded compared to heart rate. Applying filtering to this, adjusting a few heart rate samples across a patient's ICU stay would not have a large effect on any downstream task, but adjusting the results from any blood test may have a significant impact, as patients may only have one or two tests across their entire stay. Blood tests are intentionally ordered by ICU physicians either in response to other clinical data or to help clarify the meaning of it. That additional information is likely very relevant to the underlying condition affecting the patient, which is ultimately what will have an effect on most downstream tasks.

\begin{figure}[b]
    \vspace{-20pt}
    \centering
    \includegraphics[width=\columnwidth]{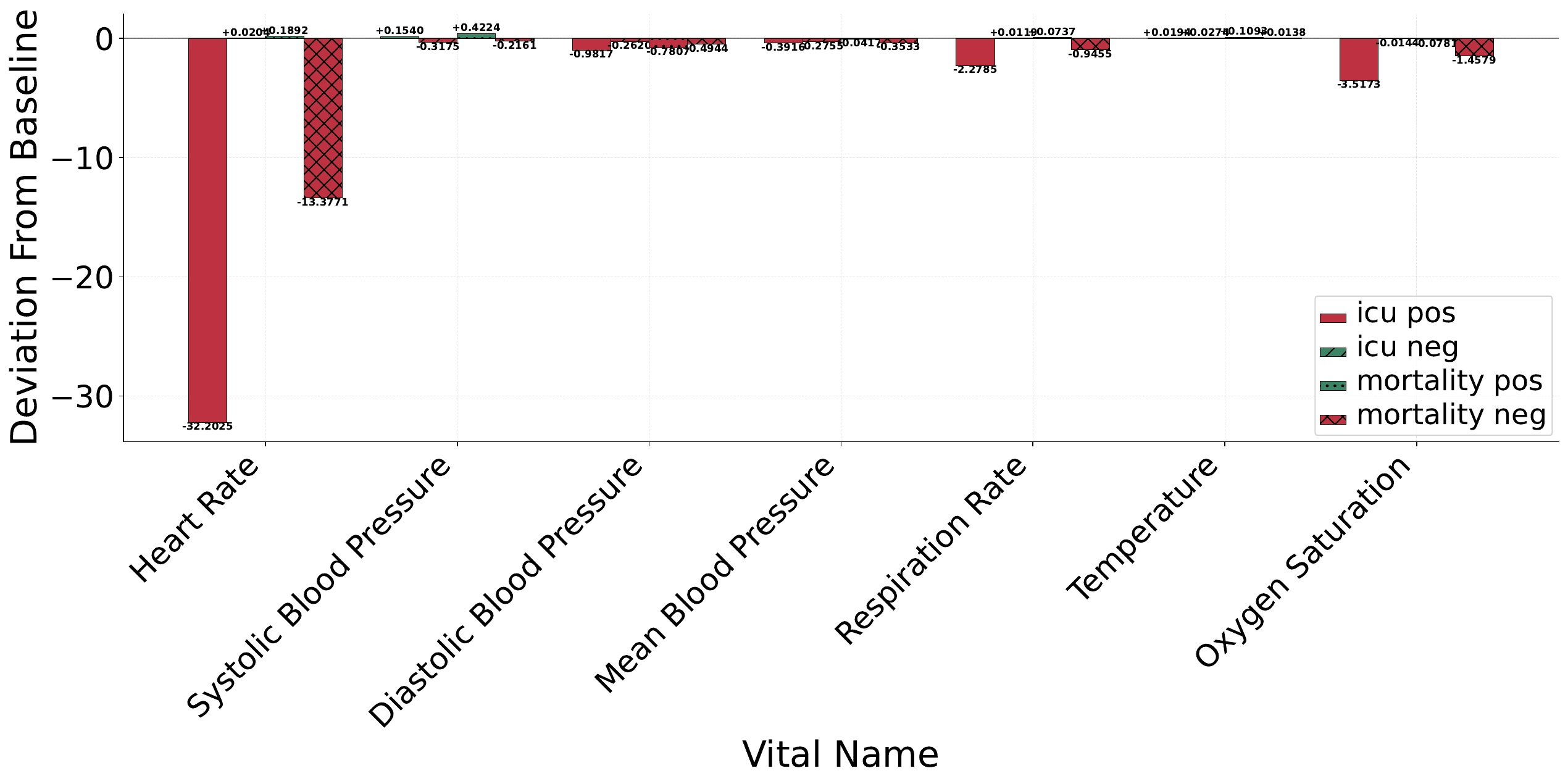}

    \caption{Centroid deviation by vital from baseline after the all-vitals outlier filter (mean aggregation). Grouped bars show the shifts for ICU  and mortality positive/negative classes.}
    \label{fig:centroid_shift_all_vitals}
    \vspace{-30pt}
\end{figure}

The effect of value-level outlier removal on the feature-space centroid is illustrated in Figure~\ref{fig:centroid_shift_all_vitals}, which shows the per-vital centroid deviation from the raw baseline for all four outcome sub-populations after the all-vitals filter is applied. As described in Table~\ref{tab:filters}, each vital-specific filter enforces a physiologically motivated inclusive range. Physiological outliers tend to cluster at one extreme of a distribution rather than being symmetric, thus removing them shifts the centroid in a direction that depends on the vital in question. Among the vitals we are looking at, the most extreme outliers are present in heart rate and oxygen saturation, which is why we see a larger shift in the centroid for these vital signs. Another thing to keep in mind is that the magnitude and direction of a centroid shift from a filter can be different between outcome groups. Such a shift calls for careful examination, as it may indicate that the underlying ML task is being made artificially easier through a fundamental change in the dataset. We examine this possibility further in the following section.

Record-level filters such as long-missing-segment, long-gap, and high-invalid-data remove entire patient records rather than individual measurements. All of these were inspired by traditional ML pre-processing techniques, where if the data does not fall into a pattern with the rest of our data, then it must be bad to train a model on. The issue with doing this is that many of our records are affected by these filters, resulting in a significant negative impact on the quality of the preprocessed data. This is because these filters target records whose monitoring was intermittent. In an ICU setting, this could mean a wide variety of things, but is likely correlated with the severity of illness, transfer events, scans or procedures that need to be performed. Therefore, by removing these records from the data set, we are shifting the more sick cohort toward patients with more complete and more regular vital-sign trajectories, effectively blurring the line between the different classes in our data set. This can be seen in Figure~\ref{fig:filter_impact_mortality_mean}, where all of our record-level filters have a very negative impact on mortality prediction. It is also worth noting that the criteria for these filters could be loosened so that they affect fewer records; however, fundamentally, any record-level filter would still be creating this problem in our data. 

\begin{figure}[b]
    \vspace{-15pt}
    \centering
    \includegraphics[width=\columnwidth]{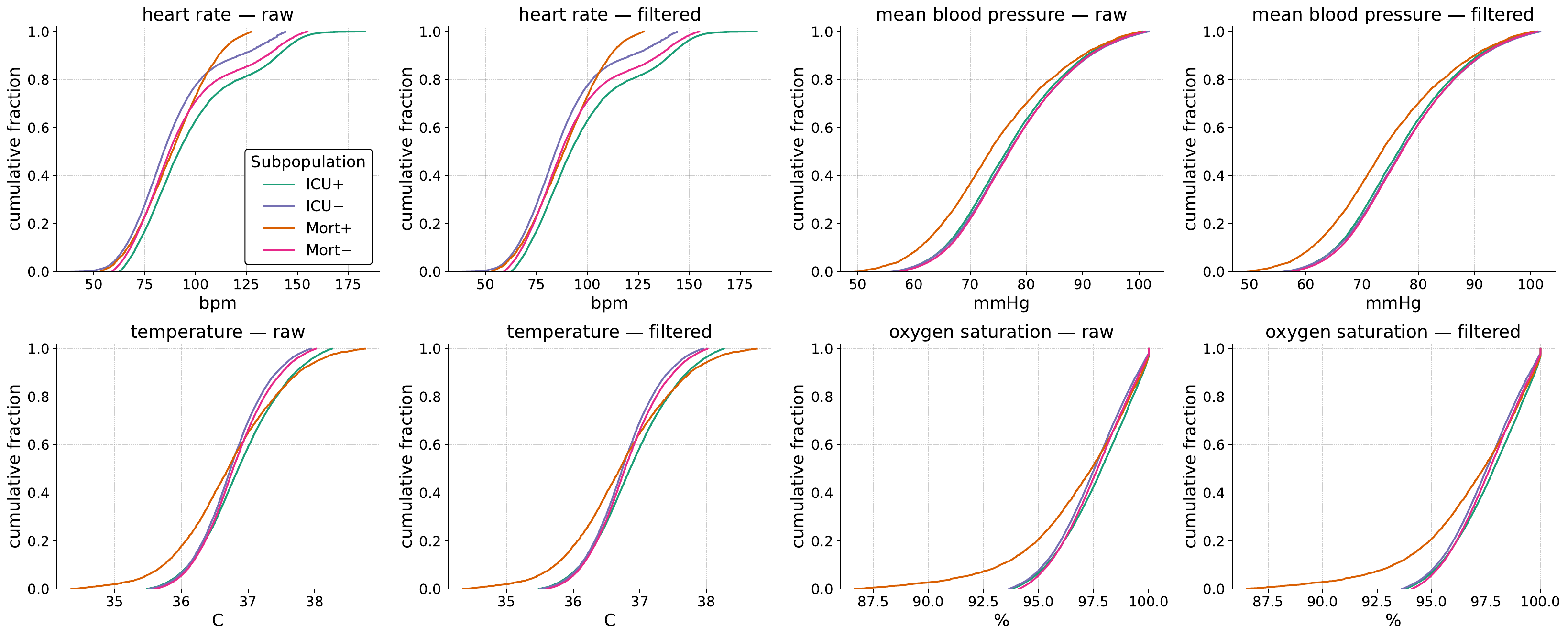}
    
    \caption{Empirical Cumulative Distribution Functions for 4 vital representatives showing the density distribution across the 4 different classes for the fill missing data filter and raw dataset.}
    \label{fig:ecdf_mean_fill_missing_data_2x4}
    \vspace{-30pt}
\end{figure}

Imputation-based filtering differs in character from both outlier removal and record gating. The fill-missing-data filter applies $k$-nearest-neighbour imputation ($k=3$) to replace \texttt{NaN} cells with values interpolated from temporal neighbours within the same record. The per-patient centroid that drives our distribution-level comparison is, however, computed by averaging only the unmasked cells of each 24-hour vital column. Cells that were \texttt{NaN} before filtering contributed nothing to the pre-filter centroid; after imputation, they carry values close to the local empirical mean. Each patient's centroid shifts by at most the imputation noise, and once aggregated across tens of thousands of records, the raw and filtered empirical cumulative distributions visible in Figure~\ref{fig:ecdf_mean_fill_missing_data_2x4} are visually indistinguishable for every vital. The lack of a shift in the centroid is not a deficiency of the filter, but a structural property of the metric. That is the centroid summary is inherently blind to whether missing cells remain masked or are imputed near the local mean. Because the imputation operates entirely on data that the centroid statistic discards, any signal it provides for downstream models will only emerge in metrics sensitive to within-record dispersion or per-vital data coverage.

\vspace{-5pt}
\subsection{Filtering Can Make Some Tasks Artificially Easier}

A natural question when evaluating data preprocessing pipelines is whether observed improvements in model performance reflect a genuine increase in the quality of the learning signal or merely an artificial simplification of the classification problem. There are several ways this can be determined, but one of the most common is to measure the deviation in test accuracy and macro F1 relative to the raw unfiltered baseline for each filter. This can be done with a wide variety of different classifiers, but we used a Random Forest classifier evaluated over four random seeds with five-fold cross-validation. To further verify our findings, we supplemented the accuracy comparisons with McNemar tests to assess whether the pattern of correct and incorrect predictions changes significantly.

\begin{figure}[b]
    \vspace{-15pt}
    \centering
    \begin{subfigure}[b]{0.49\columnwidth}
        \includegraphics[width=\textwidth]{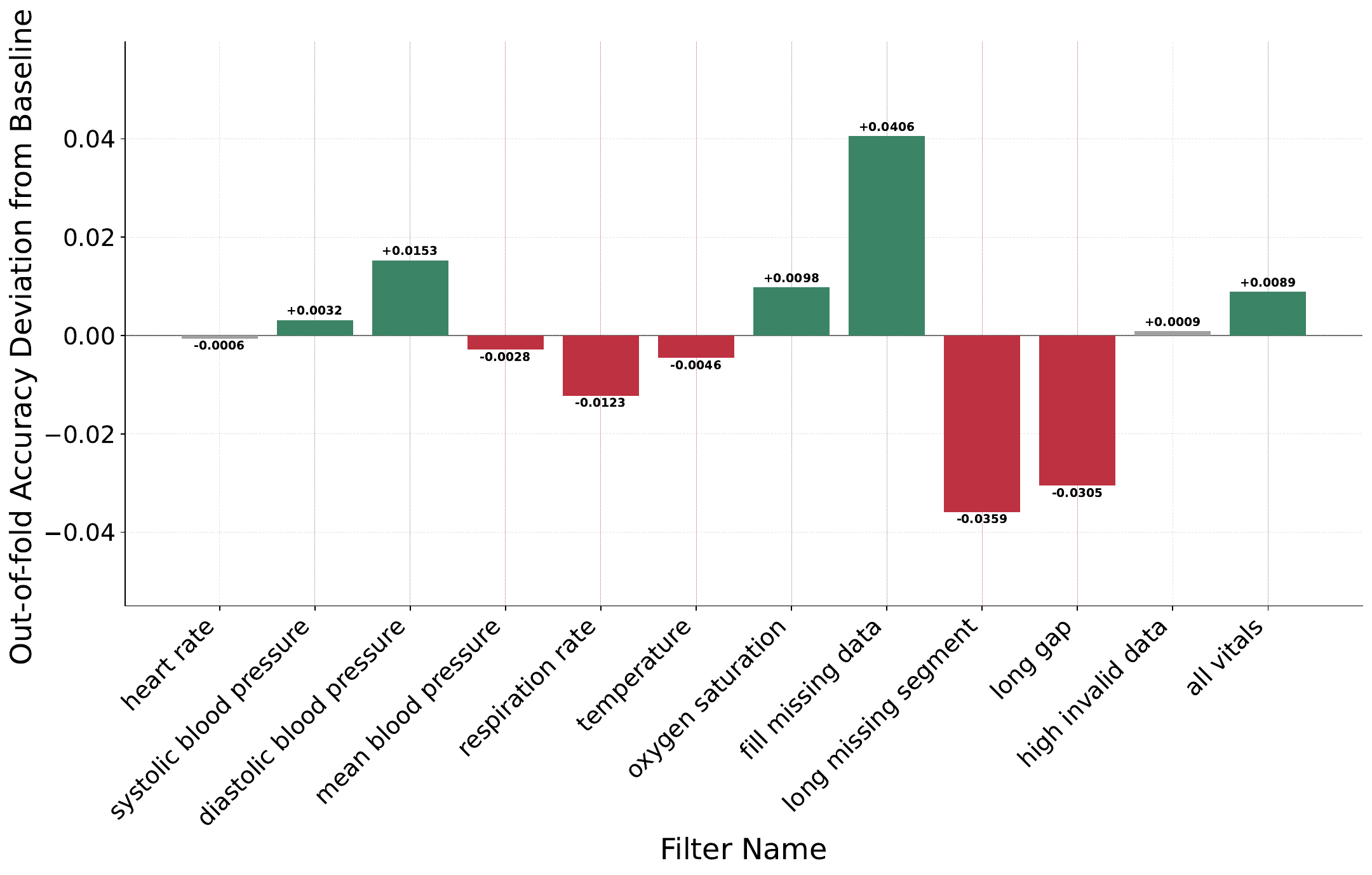}
        \caption{Test accuracy deviation}
    \end{subfigure}
    \hfill
    \begin{subfigure}[b]{0.49\columnwidth}
   
        \includegraphics[width=\textwidth]{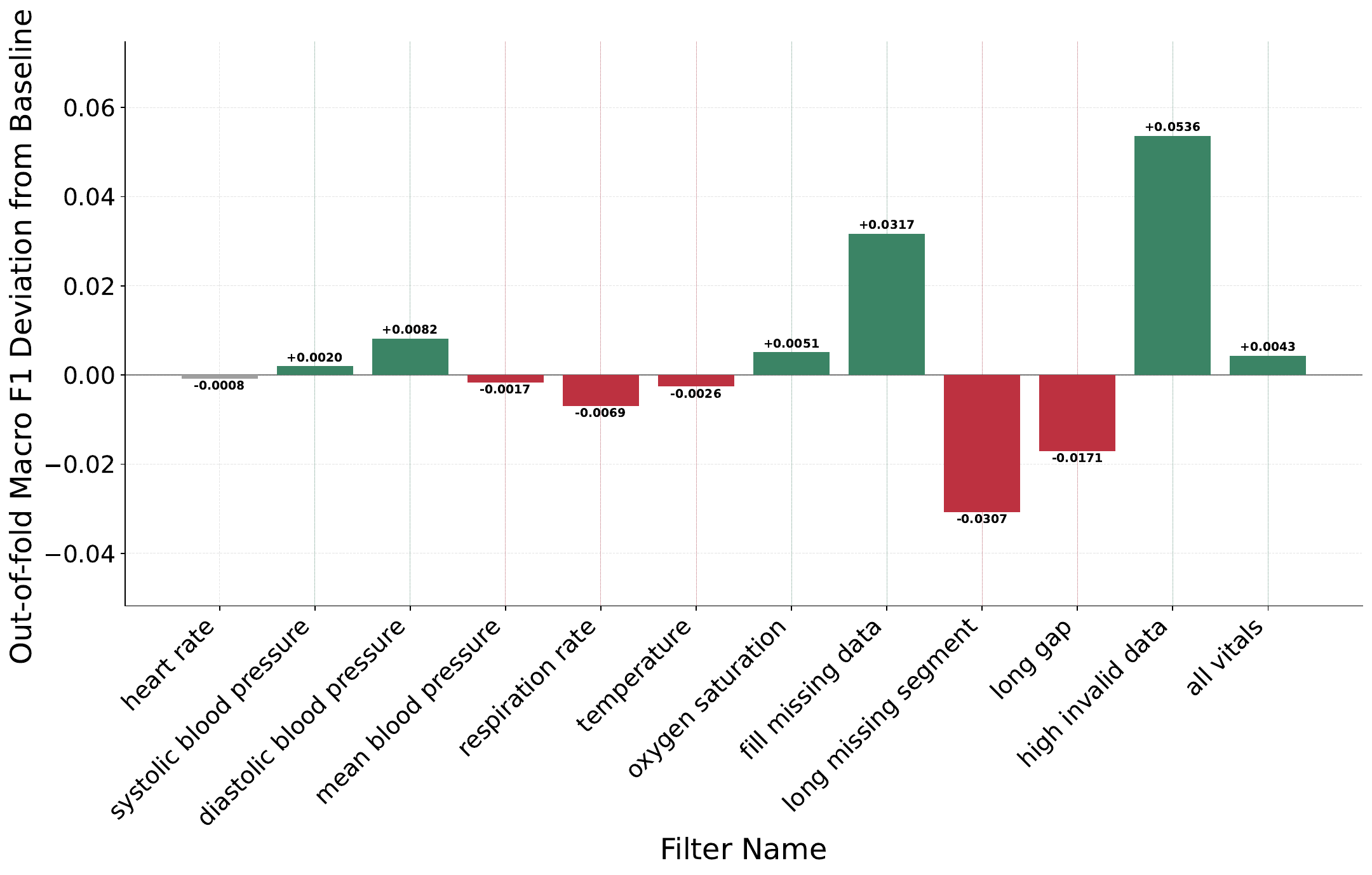}
        \caption{Test F1 deviation}
    \end{subfigure}
    \caption{Filter impact on mortality prediction under mean
             aggregation. Each bar shows the deviation in test
             accuracy (left) and macro F1 (right) relative to the raw
             unfiltered baseline, averaged over four random seeds.}
    \label{fig:filter_impact_mortality_mean}
    \vspace{-30pt}
\end{figure}

Figure~\ref{fig:filter_impact_mortality_mean} shows the per-filter accuracy and F1 deviations for mortality prediction under mean aggregation. Several of our filters produce consistent positive deviations across both metrics, suggesting that they have altered the dataset in a way that makes the classification problem easier for the model. Critically, it is important to ensure that we are still fundamentally trying to solve the same problem and not transforming our problem into a different, easier one through pre-processing. If the goal was simply to achieve the highest accuracy possible and there were no limits to what we could do with pre-processing, then we could easily create a filter that removes all of the difficult cases from our dataset. 

In the case of our experiments, just by looking at the accuracies and F1 scores from Figure~\ref{fig:filter_impact_mortality_mean}, it is clear that all the record-level filters have a big impact on the data and thus the model and downstream task. However, it is possible that a filter may not have a big impact or might have a positive impact on a downstream task while still fundamentally changing the problem at hand. One way to gain insight into when this might be happening is by looking at how a filtering operation affects the different classes that are present in the dataset. A good way to get a visual representation of this is to compute centroids for each vital sign and class in our dataset before and after we apply any filtering operation. If an operation affects different classes disproportionately without a good reason, then this is a strong indicator that there is some sort of underlying bias in the filter that should be addressed. 

For the \textbf{All Vitals} filter, looking at Figure~\ref{fig:centroid_shift_all_vitals}, we can see that the largest shift that is not common among all classes is clearly in the heart rate data. This should be investigated further to ensure that what the filter is actually changing about the data is reasonable. There are many different ways that this can be done, such as creating some sort of table like Table~\ref{tab:hr_summary_stats}. From this very simple table, we can see that our filter is not changing very many records in the data set, but the ones that we are changing have a clear effect on the max values. As all statistical measures in this table coincide with the removal of very large outliers from the data set, and because those outliers could have occurred in any of our classes, removing them is a very reasonable preprocessing step. Due to the magnitude of some of the outliers in the MIMIC-III dataset, we can determine the class certain outliers are members of just by looking at the centroids for the raw dataset, as seen in Figure~\ref{fig:centdenbase}\cite{johnson2016mimic}. Going through a process like this fully addresses why we see the centroids shift the way that they do for the heart rate in Figure~\ref{fig:centroid_shift_all_vitals} and verifies that we have not fundamentally changed anything about our problem.

\begin{table}[t]
  \centering
  \caption{Heart rate summary statistics of MIMIC-III across four levels of aggregation. Values are in beats per minute (BPM). Row~(a) is the raw dataset; row~(b)
  applies the project's physiological plausibility range; rows~(c) and~(d) are the results of applying mean aggregation over the first 24 hours of the stay before and after the heart-rate outlier filter.}
  \label{tab:hr_summary_stats}
  
  \begin{tabular}{ccccccc}
  \toprule
  Row & No.~of Records & Max & Mean & Median & Std \\
  \midrule
  a & 1{,}158{,}296 & 9{,}999{,}999 & 98.6 & 86 & 9{,}291.5 \\
  b & 1{,}157{,}789 & 459 & 90 & 86 & 24.4 \\
  c & 45{,}985 & 555{,}652 & 105.3 & 87.1 & 2{,}590.9 \\
  d & 45{,}989 & 251.5 & 93.187 & 87.1 & 25.5 \\
  \bottomrule
  \end{tabular}
  \vspace{-15pt}
\end{table}

\begin{figure}[b]
    \vspace{-20pt}
    \centering

    \includegraphics[width=\columnwidth]{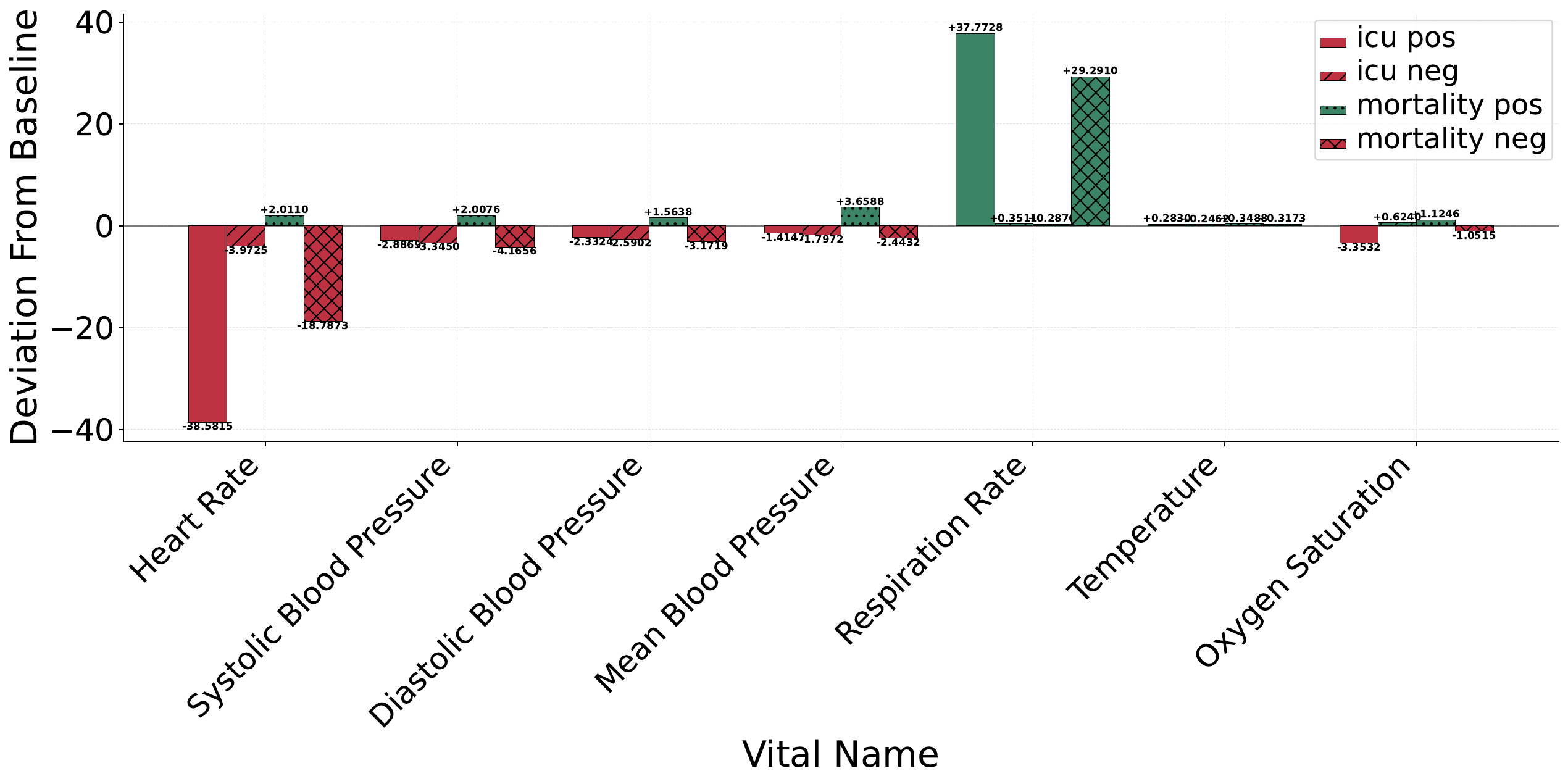}
   \caption{Centroid deviation from baseline per vital sign after the high-invalid-data filter (mean aggregation). Grouped bars show the four outcome sub-populations: ICU positive/negative and mortality positive/negative.}
    \label{fig:centroid_shift_high_invalid}
    \vspace{-38pt}
\end{figure}

On the other hand, looking at Figure~\ref{fig:centroid_shift_high_invalid} which shows the centroid shift for the high invalid data filter, we can see that there is a pretty significant shift in almost every vital sign and class across the MIMIC-III dataset. Some of these shifts may be reasonable, like we saw with the \textbf{All Vitals} filter; however, there are some things here that clearly indicate that what we are doing with the data does not make sense. For instance, we see the respiratory rate being shifted greatly upwards, which does not make a lot of sense, given there is not enough physiological wiggle room to accommodate this shift. Also, for heart rate, systolic blood pressure, diastolic blood pressure, mean blood pressure and oxygen saturation, we see that there is a consistent positive shift in the mortality positive class and a consistent negative shift in the mortality negative class. A shift pattern like this with a sufficient magnitude means either our two classes are being made more similar to one another or the separation between them is being increased across these vital signs. Which of these two things is happening depends on what the baseline centroid for each of the classes is, but we can also determine what is happening by looking at Figure~\ref{fig:filter_impact_mortality_mean}. If it were the case that the two classes were being further separated from one another, then it would make the classification problem easier, resulting in a performance increase proportional to the distance the classes are shifted apart. Indeed, Figure~\ref{fig:filter_impact_mortality_mean} confirms this increased separation: we observe a slight positive shift in accuracy, alongside a notable improvement in Macro F1 score for the high invalid data filter. Therefore, rather than bringing the classes closer together, this filter is effectively pulling the centroids further apart, artificially making the classification problem easier despite the physically improbable shifts in the underlying vitals.

\begin{figure}[b]
    \vspace{-20pt}
    \centering

    \includegraphics[width=\columnwidth]{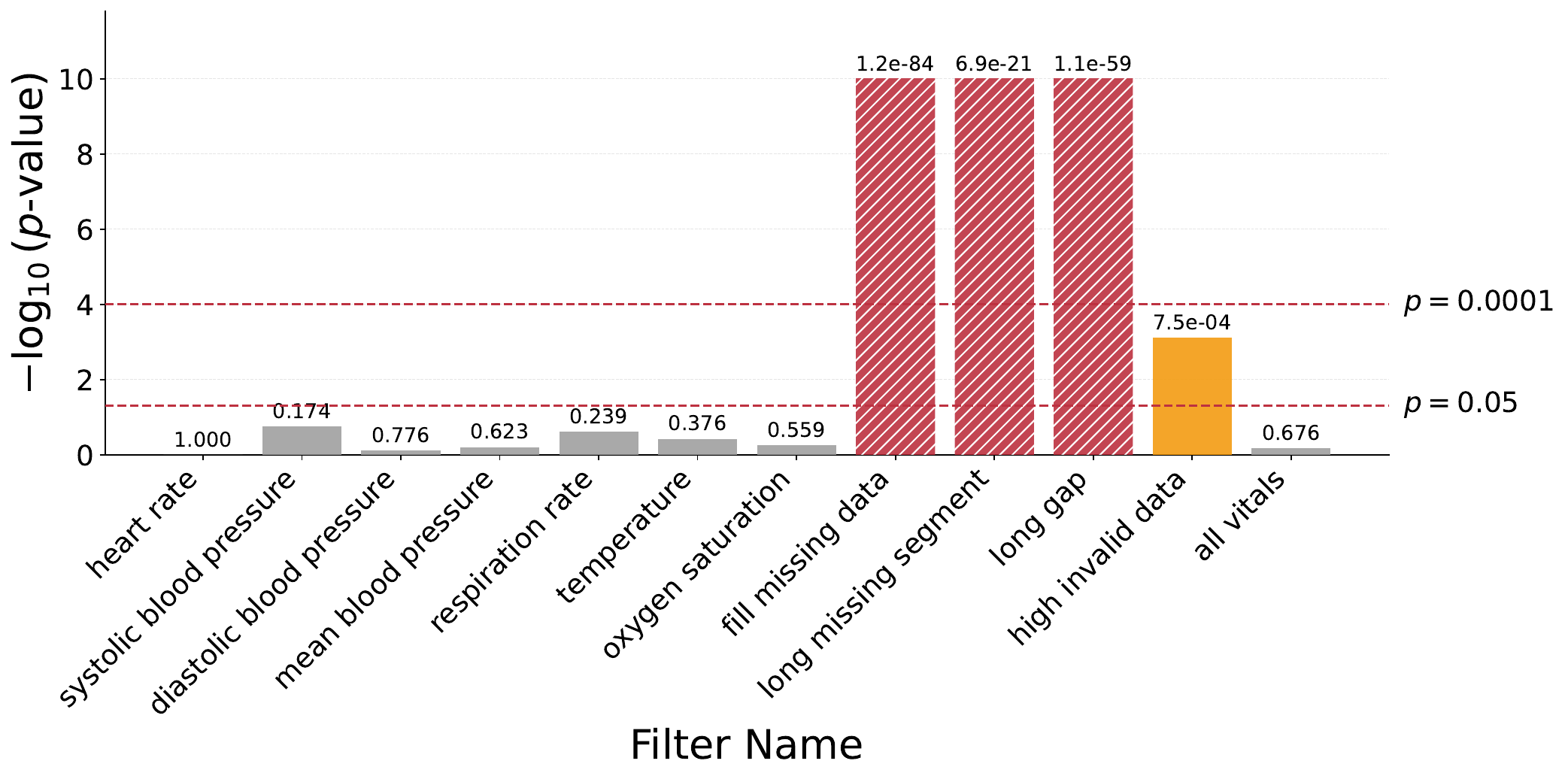}
    
    \caption{McNemar significance test for mortality prediction
             under mean aggregation. Bars show $-\log_{10}(p)$ for each
             filter versus the raw baseline. The dashed red lines mark our pre-determined
             $p = 0.05$ and $p_2=0.0001$ threasholds.}
    \label{fig:mcnemar_icu_mean}
    \vspace{-38pt}
\end{figure}

Performing a McNemar significance test is valuable because it allows us to determine whether the changes introduced by our data filters result in a statistically significant difference in downstream classifier performance. If we discover a statistically significant change in downstream task performance, then this is something that should be investigated similarly to what we did before.  Figure~\ref{fig:mcnemar_icu_mean} shows the McNemar significance test results for the mortality task. Filters whose bars exceeded the $p = 0.05$ threshold produced a statistically significant change in the model's prediction pattern.  We can also set another statistical threshold, such as $p_2 = 0.0001$, and say that any $p$-value that is smaller than this is creating so much change that we have to be fundamentally changing our problem. 

If the magnitude of a centroid shift is small, we are shifting our classes in the same direction, and a McNemar significance test yields a result that corresponds to a statistically insignificant p-value, then we can say that a given filter has not fundamentally transformed our data set in a way that impacts downstream task performance. If a McNemar significance test yields a result between the predefined $p$ and $p_2$ values, then the filter should be investigated, as it could be transforming the data set in a way that makes the underlying problem easier. Finally, if a McNemar significance test yields a smaller result than the predefined $p_2$ value, then the statistical change tells us that we are transforming the data and thus the problem we are trying to solve with the data. 

\vspace{-5pt}
\subsection{Different Tasks Respond Differently to the Same Filter}

The two prediction tasks we examined in our experiments share the same feature space and undergo identical preprocessing pipelines. Despite this, they respond differently to the same set of filters. Both tasks differ in terms of their base rates, the physiological signatures that discriminate their positive and negative classes, and in how those signatures are encoded by each aggregation method. As a result, a filter that improves one task may degrade or not affect the other.

\begin{figure}[b]
    \vspace{-5pt}
    \centering

    \includegraphics[width=0.9\columnwidth]{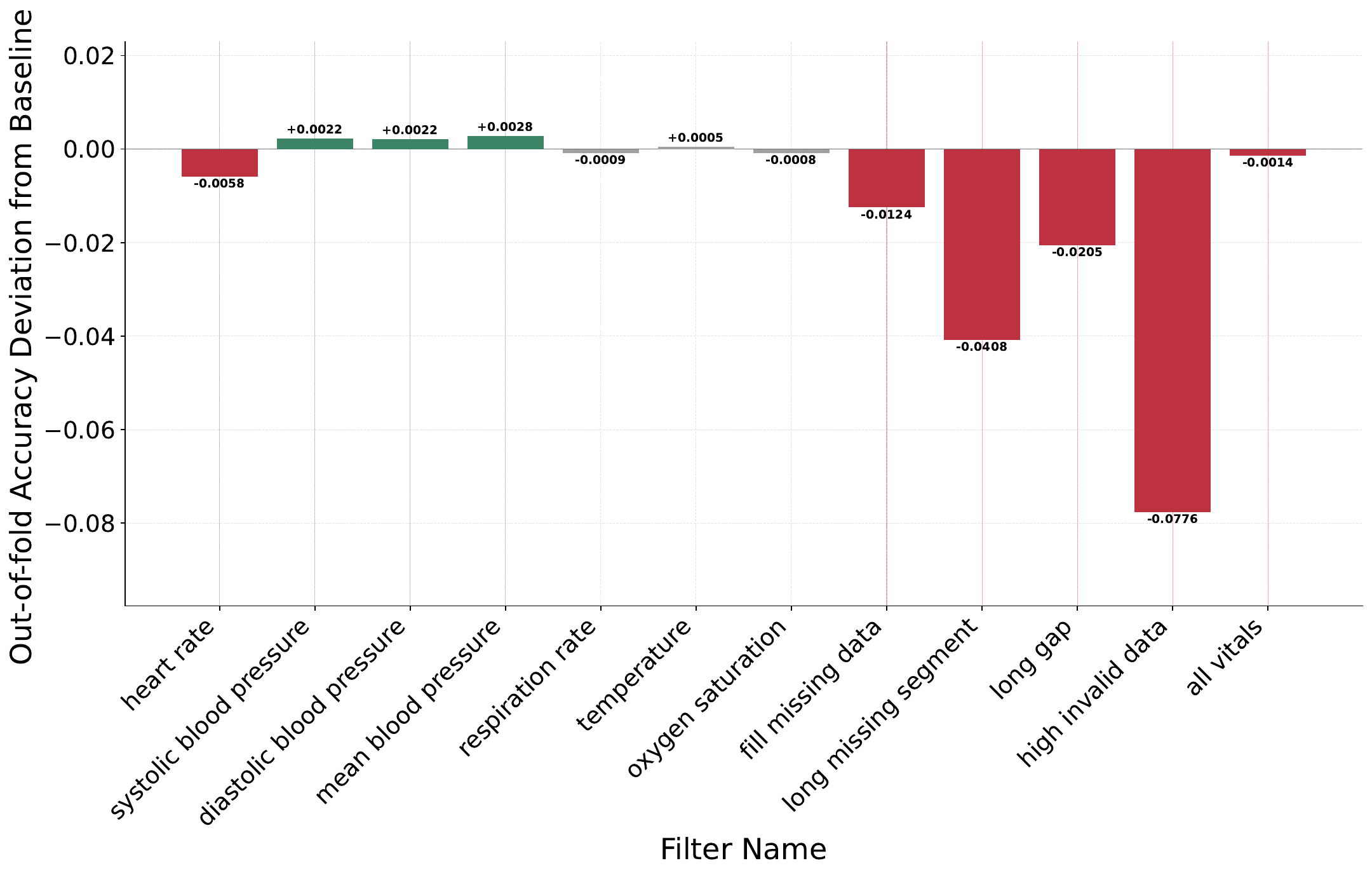}
    \caption{Filter impact on ICU prediction under mean
             aggregation. Each bar shows the deviation in test
             accuracy relative to the raw
             unfiltered baseline, averaged over four random seeds}
    \label{fig:icu_filter_impact}
    \vspace{-35pt}
\end{figure}

We can examine this result by comparing Figure~\ref{fig:filter_impact_mortality_mean} and Figure~\ref{fig:icu_filter_impact}. Both of these figures contain accuracy deviation from the raw baseline for their respective tasks under mean aggregation with the same set of filters. Looking at these graphs, we can see that several value-level filters deviate from one graph to another. This divergence is not an artifact of a particular aggregation method, as we found similar trends regardless of the method used.  The reason that this is happening stems from the records that make up each of our classes that define the downstream tasks. Our value level filters target records that contain invalid data, regardless of the class they are a member of. What this means is that if the records with invalid data were making the baseline problem easier by adding additional separation between the classes, then by removing the outliers through our value level filters, we are fundamentally uncovering the difficulty of the problem at hand. The opposite is also true: if our outliers were making the problem more difficult, then removing them is letting us tackle the real problem at hand. In a real clinical setting, we do not want the model ever to make predictions based on physiologically impossible values. Consequently, even if the \textbf{All Vitals} filter with mean aggregation yields a slight performance drop for ICU prediction compared to the baseline, it remains preferable in clinical contexts. As AI approaches integration into live medical environments, human explainability is paramount. Models must be trained exclusively on physiologically valid data to prevent anomalous artifacts from improperly skewing clinical decisions.

\vspace{-5pt}
\subsection{Linking Data Changes to Task Difficulty}

The above analyses suggest that our visualization toolbox can serve as a supporting tool for interpretation rather than a definitive test. A researcher can view the per-class centroid deviations in Figures~\ref{fig:centroid_shift_all_vitals} and~\ref{fig:centroid_shift_high_invalid} alongside Table~\ref{tab:hr_summary_stats}, the downstream accuracy and McNemar results in Figures~\ref{fig:filter_impact_mortality_mean}, ~\ref{fig:mcnemar_icu_mean} and~\ref{fig:icu_filter_impact} to relate a filter's effect on the data to its effect on task difficulty. In our examples, this offered a useful signal of when a filter appears to remove artifacts, as with the \textbf{All Vitals} filter, and when it may be altering the underlying problem, as with the high-invalid-data filter. This helps researchers assess whether a filtered dataset still reflects the real clinical task rather than an artificially simplified version of it.

\vspace{-5pt}
\subsection{Comparison with MIMIC-Extract}
To situate our pipeline relative to existing work on MIMIC-III~\cite{johnson2016mimic}, we compared the downstream performance of data produced by our pipeline against that of the widely used MIMIC-Extract pipeline~\cite{wang2020mimic}. Run in their standard configurations without additional filtering, the two pipelines yielded comparable downstream performance. A broader comparison across multiple filtering configurations proved difficult to interpret, as the two pipelines expose substantially different filtering options, which makes a direct comparison impractical. Overall, both pipelines produce high-quality data; our pipeline is better suited to studies where understanding and auditing the effect of each filtering choice is a priority, whereas MIMIC-Extract offers a broader set of pre-built features for studies that need extensive preprocessing out of the box. We hope to publish the results of this comparison in another paper; however, in the meantime, all figures and documentation for this comparison are available in our repository.

\vspace{-5pt}
\section{Implications}

Our results suggest that filtering is not a neutral cleaning step but an active modeling choice that reshapes both the data distribution and the difficulty of the downstream task. 

\vspace{-5pt}

\subsection{Implications for Algorithm Comparison}
Because a single filter can shift class centroids and significantly change a model's prediction pattern (Figures \ref{fig:centroid_shift_high_invalid} and \ref{fig:mcnemar_icu_mean}), two studies applying different filters to the same dataset may be solving problems of different difficulty. This effect is task-dependent: the same filter can help one task while degrading another (Figures \ref{fig:filter_impact_mortality_mean} and \ref{fig:icu_filter_impact}). Reported performance gaps may therefore reflect preprocessing rather than model quality, so cross-study comparisons should account for filtering choices, not only model design.

\vspace{-5pt}

\subsection{Implications for Clinical Usefulness}

High performance on heavily filtered benchmarks does not guarantee clinical readiness. While standard pipelines often treat missing data as random noise, electronic health records typically exhibit Missing Not At Random (MNAR) patterns, where data absence reflects clinical decisions or patient acuity \cite{tan2022}. Because observation frequency correlates directly with disease burden \cite{li2021}, aggressive complete-case filtering introduces severe selection bias by omitting these informative gaps \cite{beaulieu2017, gao2025}. In our analysis, record-level filters disproportionately excluded intermittently monitored patients, thereby obscuring outcome class boundaries and degrading mortality prediction (Figure~\ref{fig:filter_impact_mortality_mean}). Therefore, we advocate for filtering strategies that prioritize physiological validity over marginal gains in benchmark performance.

\vspace{-5pt}

\subsection{Limitations}
A primary limitation of this work is its reliance on a flattened hourly aggregation architecture. While standard for benchmarks, hourly flattening smooths away sub-hourly physiological volatility and informative missingness (MNAR) patterns that are deeply prognostic of acute patient deterioration. To bridge this clinical gap for high-dependency environments, future architectural phases should explore pipeline designs that are capable of preserving and evaluating raw physiological volatility and sampling dynamics directly.

\section{Conclusion}
 
Filtering is not merely a technical preprocessing step; it can change the meaning, difficulty, and interpretability of clinical machine learning tasks. Using an explainable filtering toolbox, we showed that individual filtering choices measurably reshape the data distribution and can make downstream tasks artificially easier or harder. We introduced a diagnostic method  combining centroid-deviation analysis with McNemar testing to help detect when a filter has altered the underlying problem. We acknowledge that our study is limited to the MIMIC~\cite{johnson2016mimic, johnson2023mimic} and eICU~\cite{pollard2018eicu} databases, seven vital signs, two tasks, and a single classifier, and we leave extending it to other datasets, tasks, and models as future work.



\IEEEtriggeratref{12}


\bibliographystyle{IEEEtran}
\bibliography{references}

\end{document}